\documentclass[conference]{IEEEtran}
\IEEEoverridecommandlockouts

\usepackage{amsmath,amssymb}
\usepackage{amsfonts}
\usepackage{graphicx}
\usepackage{hyperref}
\usepackage{listings}
\usepackage{xcolor}
\usepackage{url}
\usepackage{cite}
\usepackage{booktabs}
\usepackage{multirow}
\usepackage{orcidlink}
\usepackage{colortbl}
\usepackage{tabularx}

\definecolor{rowgray}{RGB}{245,245,245}
\definecolor{codegreen}{rgb}{0.0,0.4,0.0}
\definecolor{codegray}{rgb}{0.4,0.4,0.4}
\definecolor{codeblue}{rgb}{0.0,0.2,0.6}

\lstdefinelanguage{turtle}{
  morekeywords={@prefix, a, sh:NodeShape, sh:property, sh:targetClass,
                sh:minCount, sh:datatype, sh:path, sh:message, sh:sparql,
                sh:SPARQLConstraint, sh:select, FILTER, BIND, SELECT,
                WHERE, NOT, EXISTS, ABS, IF},
  sensitive=true,
  morecomment=[l]{\#},
  morestring=[b]"
}

\title{Ontological Knowledge Blocks: Executable Compliance and Profile-Based Validation for Trustworthy AI Systems}

\author{
\IEEEauthorblockN{Aasish Kumar Sharma\,\orcidlink{0000-0002-7514-2340} and Julian M. Kunkel\,\orcidlink{0000-0002-6915-1179}}
\IEEEauthorblockA{Institute of Computer Science, Georg-August-Universität Göttingen / GWDG, Germany\\
\{aasish.sharma, julian.kunkel\}@uni-goettingen.de}
}

\begin{document}
\maketitle

\begin{abstract}
AI-enabled services deployed in critical digital infrastructure are subject to governance obligations spanning transparency, accountability, fairness, and traceability. Compliance today remains documentation-centric: obligations are described in prose, audits rely on static checklists, and verification depends on manual review. Such approaches do not scale to automated AI systems. This paper introduces Ontological Knowledge Blocks (OKBs), a programmable governance infrastructure that compiles regulatory obligations into machine-checkable constraints over structured evidence graphs. We formalize an OKB as a 5-tuple $\langle O, C, V, E, P\rangle$ that binds normative obligations to an RDF/OWL concept schema, executable SHACL validation rules, explicit evidence requirements, and PROV-O provenance links. A deterministic regulatory compiler translates structured Intermediate Representation (IR) records into composable KB modules, enabling profile-based governance reconfiguration without modifying service code. We implement two prototypes and evaluate them in an AI-assisted HPC resource allocation scenario across 24 validation runs and four governance profiles. Results demonstrate profile-sensitive validation, strictly additive violation accumulation (Accountability: 1, Fairness: 2, Combined: 3 violations across failure cases), SHACL validation latency between 12.6\,ms and 100.3\,ms, and profile equivalence testing confirming Combined as the strictly most comprehensive profile, subsuming both Accountability and Fairness with no equivalence pairs. All artefacts are released as open source.
\end{abstract}


\section{Introduction}

Responsible AI governance is crossing a threshold. What was once a field of voluntary guidelines~\cite{JobinIencaVayena2019,Hagendorff2020} is becoming a landscape of enforceable, operational obligations. The EU AI Act requires high-risk systems to maintain technical documentation, record decisions with timestamps, provide transparency notices, and implement human oversight mechanisms~\cite{EUAIAct2024}. The NIST AI Risk Management Framework (RMF) structures these concerns across the full AI lifecycle~\cite{NISTAI100-1}. The practical question is, \emph{not whether governance obligations exist, they do, but whether they can be \textbf{checked automatically} against the evidence an AI system actually produces?}

The honest answer, today, is ``\emph{no}''. Compliance practice still relies on narrative documentation compiled for periodic audits. That approach has a structural flaw: it records \emph{claims} rather than verifying \emph{evidence}. A document asserting that ``logging is in place'' cannot confirm that individual decisions carry correctly typed timestamps, that provenance links reach the model artifact that generated the decision, or that resource allocations satisfy a declared fairness threshold.

\emph{Problem:} To our knowledge, no widely adopted, standards-aligned, composable mechanism currently (a) makes governance obligations explicit and human-reviewable, (b) compiles them deterministically into executable checks, and (c) validates those checks continuously over evidence emitted by AI services. While individual components exist (ontology-based compliance models~\cite{GallinaJSEP2024,GallinaICSR2024}, SHACL-based compliance checking~\cite{AnimSHACL2024}, and policy-as-code engines~\cite{OPA}), no existing system, to our knowledge, integrates semantic evidence representation, deterministic obligation compilation, and executable profile-based validation into a single pipeline.

\emph{Approach:} We propose \emph{Ontological Knowledge Blocks} (OKBs), a programmable compliance substrate built on W3C standards. Each OKB encodes governance obligations as an RDF/OWL concept schema paired with SHACL constraint shapes that execute over an \emph{evidence graph}. A regulatory compiler translates human-reviewed IR records into Knowledge Block (KB) modules deterministically. Governance profiles select and compose KBs without touching service code.

\emph{Contributions:} (i)~\emph{Formalization:} an OKB model KB\,$= \langle O, C, V, E, P\rangle$ with a compliance decision procedure and profile composition algebra. (ii)~\emph{Compiler:} a deterministic IR-to-SHACL compiler. (iii)~\emph{Prototype:} open-source reference implementations covering logging, transparency, fairness, and provenance. (iv)~\emph{Evaluation:} validation of profile sensitivity, refinement relations, and sub-100\,ms latency; Combined confirmed as strictly the most comprehensive profile.

\emph{Scope:} This work does not automate legal interpretation. Normative decisions remain with human governance owners. Once an obligation is captured in IR and reviewed, compilation and checking are deterministic and repeatable.

\section{Background}
\label{sec:background}

\emph{RDF and OWL:} The Resource Description Framework (RDF)~\cite{RDF11} defines a graph data model for representing entities and relationships as subject--predicate--object triples. The Web Ontology Language (OWL) extends RDF with formal class hierarchies. Together they provide a standards-based substrate for structured evidence representation.

\emph{SHACL:} The Shapes Constraint Language (SHACL)~\cite{SHACL2017} provides machine-checkable constraint shapes over RDF graphs. SHACL Core supports structural and datatype validation; SHACL-SPARQL extends this with query-based constraints enabling quantitative checks such as threshold comparisons.

\emph{PROV-O:} The PROV Ontology (PROV-O)~\cite{PROVO2013} expresses provenance relationships between entities, activities, and agents. In this work, PROV-O links AI decisions to the model artifacts and activities that generated them, enabling traceability audits required by the EU AI Act and NIST AI RMF.

\section{Related Work}

\subsection{Governance Frameworks and Documentation Artefacts}
The NIST AI RMF~\cite{NISTAI100-1} structures governance across the AI lifecycle. Model Cards~\cite{MitchellModelCards2019} and Datasheets for Datasets~\cite{GebruDatasheets2021} standardise how AI systems are described. These artefacts support human inspection but contain nothing executable and carry no mechanism for profile-based governance variation.

\subsection{Policy-as-Code and Automated Enforcement}
Policy-as-code systems such as OPA~\cite{OPA} enforce declarative rules at infrastructure boundaries. They are effective for access control but operate over syntactic system state rather than semantically linked evidence graphs. Provenance-aware claims about AI decisions fall outside their native data model.

\subsection{Ontology-Based Compliance Automation}
Recent work applies ontological methods to regulatory compliance. Gallina~et~al.~\cite{GallinaJSEP2024} propose an ontology for process compliance with machinery legislation; subsequent work extends this to compliance-aware system change management~\cite{GallinaEuroSPI2025} and ontology-based product evolution in regulated industries~\cite{GallinaICSR2024}. Anim~et~al.~\cite{AnimSHACL2024} demonstrate SHACL-based automated compliance checking over RDF data, confirming that SHACL-SPARQL constraints are necessary for temporal and aggregate legal requirements. These contributions advance compliance modelling but do not address composing executable governance profiles over runtime AI evidence graphs, nor provide a deterministic compilation pipeline from human-reviewed obligations to SHACL shapes.

\subsection{Positioning}
Table~\ref{tab:related} characterises OKBs against prior approaches. OKBs integrate executable validation, composable modularity, cross-profile governance variation, and evidence grounding: a combination that, to our knowledge, has not been previously demonstrated.

\begin{table}[t]
\centering
\caption{Comparison of compliance approaches. $\checkmark$\,=\,Yes; $\times$\,=\,No; $\sim$\,=\,Partial.}
\label{tab:related}
\scriptsize
\begin{tabular}{lcccc}
\toprule
\textbf{Approach} & \textbf{Exec.} & \textbf{Compos.} & \textbf{Cross-} & \textbf{Evidence} \\
 & & & \textbf{profile} & \textbf{grounded} \\
\midrule
NIST AI RMF & $\times$ & $\times$ & $\times$ & $\times$ \\
Model Cards & $\times$ & $\times$ & $\times$ & $\times$ \\
Policy-as-code (OPA) & $\checkmark$ & $\sim$ & $\times$ & $\sim$ \\
Compl.\ ontologies~\cite{GallinaJSEP2024,AnimSHACL2024} & $\sim$ & $\times$ & $\times$ & $\times$ \\
This work (OKBs) & $\checkmark$ & $\checkmark$ & $\checkmark$ & $\checkmark$ \\
\bottomrule
\end{tabular}
\end{table}

\section{Methodology}
\label{sec:methodology}

The methodology translates governance obligations into machine-checkable infrastructure through four layers, separating normative interpretation (human responsibility) from constraint execution (deterministic machine process).

\emph{Layer~1 - Obligation Representation:} Obligations are captured as structured \emph{Regulatory IR} records in YAML, specifying target concept, required relation/attribute, constraint type (structural, datatype, or quantitative threshold), parameters, and severity. IR records are authored and approved by human compliance engineers before compilation, preserving normative authority.

\emph{Layer~2 - Deterministic Compilation:} A compiler $\mathcal{C}$ maps each IR record to a SHACL node shape within a KB module:
\begin{equation}
  \mathcal{C}: \mathrm{IR} \;\mapsto\; \mathrm{KB} = \langle O, C, V, E, P\rangle.
\end{equation}
$O$ is the obligation set; $C$ the RDF/OWL concept schema; $V$ the SHACL shape set; $E$ the required evidence artifacts; $P$ the PROV-O provenance links. The mapping is deterministic: the same IR always produces the same shapes.

\emph{Layer~3 - Evidence Graph Construction:} Each AI service decision is accompanied by an RDF evidence graph linking the decision to provenance artifacts, usage logs, explanation references, and quantitative measurements. The evidence graph is emitted by the service itself, making compliance checking concurrent with operation.

\emph{Layer~4 - Profile Composition and Validation:} A governance \emph{profile} selects KB modules. Composition is defined by set union:
\begin{equation}
  \mathrm{KB}_a \oplus \mathrm{KB}_b := \langle O_a \cup O_b,\; C_a \cup C_b,\; V_a \cup V_b,\; E_a \cup E_b,\; P_a \cup P_b \rangle.
\end{equation}
The SHACL engine evaluates all shapes over the evidence graph $G_S$:
\begin{multline}
  \mathrm{validate}(G_S,\, \mathrm{KB}_a \oplus \mathrm{KB}_b) \rightarrow\\
  \langle \mathit{conforms},\, \mathit{violations},\, G_{\mathit{report}} \rangle,
\end{multline}
where $G_{\mathit{report}}$ is a machine-readable SHACL report. Profile $P_1$ \emph{refines} $P_2$ ($P_1 \sqsubseteq P_2$) if every violation detected by $P_2$ is also detected by $P_1$; equivalence holds when both directions hold. When validation fails, the report identifies which shape fired, on which node, with what message, closing the feedback loop for remediation or audit.

\section{Framework Architecture}
\label{sec:architecture}

\begin{figure}[t]
\centering
\includegraphics[width=0.3\textwidth]{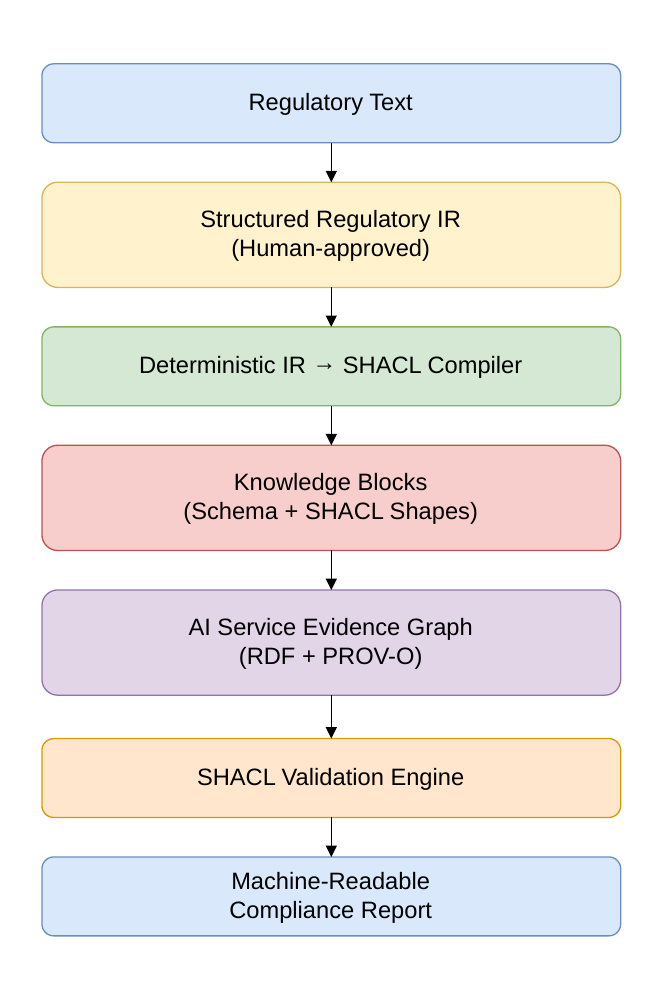}
\caption{Executable governance pipeline: obligations are compiled into KB modules and validated against evidence graphs emitted per AI decision.}
\label{fig:compiler_architecture}
\end{figure}

\begin{figure}[t]
\centering
\includegraphics[width=0.98\columnwidth]{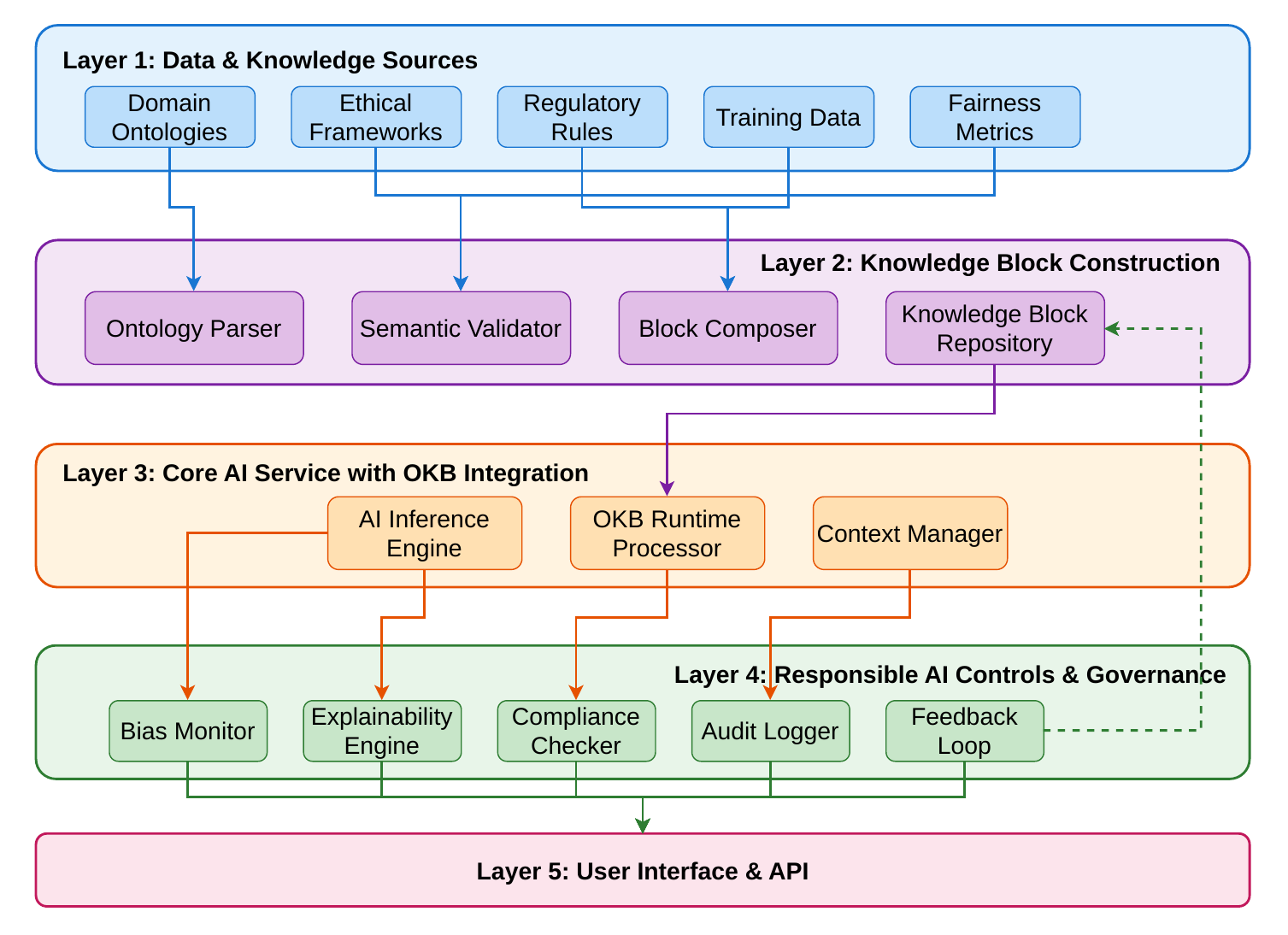}
\caption{OKB system architecture: obligations represented in IR, compiled into KB modules, executed over RDF evidence graphs at runtime.}
\label{fig:architecture}
\end{figure}

\begin{figure}[t]
\centering
\includegraphics[width=0.95\columnwidth]{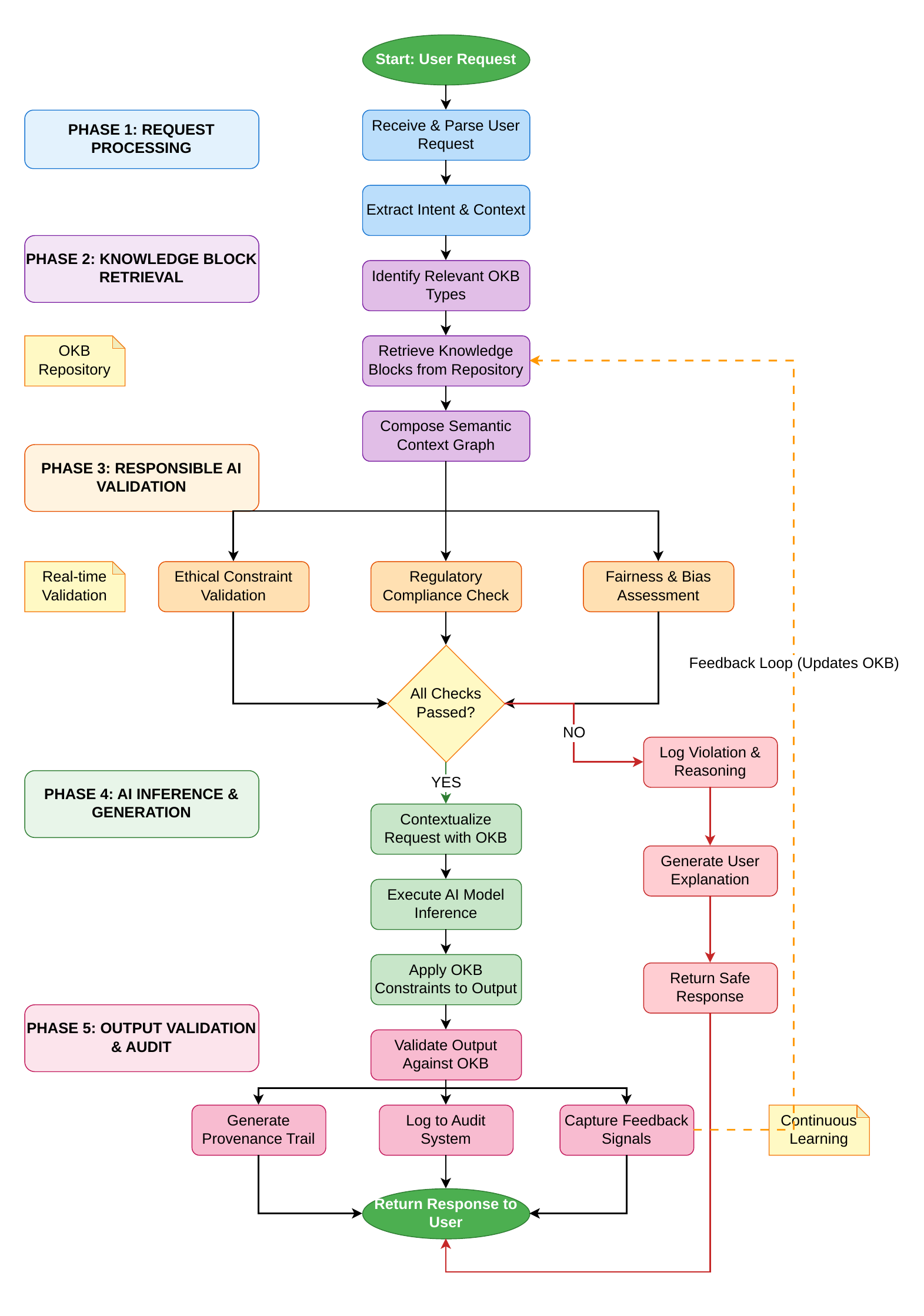}
\caption{Operational process: the AI service emits an evidence graph per decision; the governance engine selects the active profile, merges SHACL shapes, and validates. The report feeds the audit trail.}
\label{fig:process}
\end{figure}

Figure~\ref{fig:compiler_architecture} shows the governance pipeline. Figure~\ref{fig:architecture} shows the system architecture, and Figure~\ref{fig:process} shows the operational flow from evidence capture through profile selection to validation and reporting.

\section{Prototype: Scenario, Blocks, and Cases}
\label{sec:prototype}

\subsection{The HPC Resource Allocation Scenario}
We instantiate the framework on an AI-assisted GPU-hour scheduler distributing resources across research groups. HPC schedulers make high-throughput decisions affecting measurable quantities, operate across regulatory boundaries, and their fairness properties can be quantified precisely.

\subsection{Knowledge Blocks}
The \emph{Accountability block}~(A) comprises five SHACL shapes (A1--A5) enforcing logging, provenance linkage, and model traceability. The \emph{Fairness/Transparency block}~(B) comprises five shapes (B1--B5) enforcing explanation presence, fairness threshold declaration, per-group allocations, and the quantitative disparity check shown in Listing~\ref{lst:b5shape}:

\begin{lstlisting}[caption={B5Shape: SPARQL constraint for quantitative fairness.},label={lst:b5shape}]
ex:B5Shape a sh:NodeShape ;
  sh:message "Fairness disparity exceeds threshold." ;
  sh:targetClass ex:Decision ;
  sh:sparql [ a sh:SPARQLConstraint ;
    sh:select """SELECT $this WHERE {
      $this ex:allocatedGPUHoursGroupA ?a ;
            ex:allocatedGPUHoursGroupB ?b ;
            ex:fairnessThreshold       ?t .
      BIND(IF(?a > ?b, ?a, ?b) AS ?mx)
      BIND(IF(?mx = 0, 0, (ABS(?a - ?b) / ?mx)) AS ?ratio)
      FILTER(?ratio > ?t)
    }""" ] .
\end{lstlisting}

The \emph{Combined block} is the set union of A and B (all ten shapes).

\subsection{Profiles and Evidence Cases}
The \emph{OKB prototype} uses four jurisdiction-oriented profiles: EU, US, China, and EU+Fairness. The \emph{compiler prototype} uses three capability-oriented profiles: Accountability, Fairness, and Combined. Evidence cases isolate specific governance failures (details in Appendix~\ref{app:cases}).

\subsection{Implementation}
Both prototypes use Python with \texttt{rdflib}~7.6.0 and \texttt{pyshacl}~0.31.0 with RDFS inference enabled and advanced SPARQL constraint evaluation. All policy artefacts (IR obligation records, SHACL shapes, validation profiles, and evidence graphs) are version-controlled with DVC~\cite{DVC} alongside Git. DVC checksums detect any modification to policy files, providing a tamper-evident audit trail as part of the compliance security framework.

\section{Evaluation}
\label{sec:evaluation}

\subsection{Experimental Setup}
Experiment~1: 3 cases $\times$ 4 profiles = 12 runs (OKB prototype). Experiment~2: 4 cases $\times$ 3 profiles = 12 runs (compiler prototype). Total: 24 validation runs.

\subsection{Profile Sensitivity}

\begin{table}[t]
\centering
\caption{OKB prototype: conformance (\checkmark/$\times$) and violation counts.}
\label{tab:conformance_exp1}
\scriptsize
\begin{tabular}{lcccc}
\toprule
\textbf{Evidence case} & \textbf{EU} & \textbf{US} & \textbf{China} & \textbf{EU+Fairness} \\
\midrule
Case Conform & $\checkmark$\,(0) & $\checkmark$\,(0) & $\checkmark$\,(0) & $\checkmark$\,(0) \\
Case Profile & $\times$\,(1) & $\checkmark$\,(0) & $\checkmark$\,(0) & $\times$\,(1) \\
Case Violate & $\times$\,(2) & $\times$\,(1) & $\times$\,(1) & $\times$\,(3) \\
\bottomrule
\end{tabular}
\end{table}

Table~\ref{tab:conformance_exp1}: Case Conform passes all profiles. Case Profile passes US and China but fails EU and EU+Fairness with exactly one violation (missing explanation), demonstrating that the same evidence produces different verdicts via profile selection. Case Violate accumulates violations monotonically (1, 1, 2, 3).

\begin{table}[t]
\centering
\caption{Compiler prototype: conformance (\checkmark/$\times$) and violation counts.}
\label{tab:conformance_exp2}
\scriptsize
\begin{tabular}{lccc}
\toprule
\textbf{Evidence case} & \textbf{Acct} & \textbf{Fairness} & \textbf{Combined} \\
\midrule
Case Conform & $\checkmark$\,(0) & $\checkmark$\,(0) & $\checkmark$\,(0) \\
Case Missing Explanation & $\checkmark$\,(0) & $\times$\,(1) & $\times$\,(1) \\
Case Disparity Exceeds & $\checkmark$\,(0) & $\times$\,(1) & $\times$\,(1) \\
Case Missing Model Artifact & $\times$\,(1) & $\checkmark$\,(0) & $\times$\,(1) \\
\bottomrule
\end{tabular}
\end{table}

Table~\ref{tab:conformance_exp2}: Atomic failure cases are detected precisely by the profiles encoding the corresponding obligations. Case~C (Missing Model Artifact) fails Accountability and Combined, as expected: A5Shape correctly targets provenance activity nodes co-typed as \texttt{ex:Activity}. Violation accumulation is strictly additive: Accountability detects 1 failure case, Fairness detects 2, Combined detects all 3.

\subsection{Profile Refinement and Design Insight}

\begin{table}[t]
\centering
\caption{Profile refinement relations. $\checkmark$\,=\,Holds; $\times$\,=\,Does not hold.}
\label{tab:refinement}
\scriptsize
\begin{tabular}{lcl}
\toprule
\textbf{Relation} & \textbf{Holds?} & \textbf{Implication} \\
\midrule
Fairness $\sqsubseteq$ Accountability & $\times$ & Fairness misses provenance (A5) \\
Combined $\sqsubseteq$ Accountability & $\checkmark$ & Combined detects all Acct. violations \\
Combined $\sqsubseteq$ Fairness & $\checkmark$ & Combined detects all Fairness violations \\
Fairness $\sqsubseteq$ Combined & $\times$ & Combined additionally detects Case~C \\
Accountability $\sqsubseteq$ Fairness & $\times$ & Misses explanation/disparity \\
Accountability $\sqsubseteq$ Combined & $\times$ & Misses fairness obligations \\
\midrule
\multicolumn{3}{l}{\textit{Combined is strictly the most comprehensive profile}} \\
\bottomrule
\end{tabular}
\end{table}

Table~\ref{tab:refinement} shows Combined is \emph{strictly} the most comprehensive profile: it subsumes both Accountability and Fairness but is subsumed by neither. No two profiles are equivalent; each encodes a distinct obligation scope. Profile equivalence testing is a precise structural diagnostic: two profiles are equivalent if and only if they fire on exactly the same evidence graphs, which the refinement algebra detects without manual inspection.

\subsection{Validation Performance}

\begin{table}[t]
\centering
\caption{Validation latency (ms) across evidence cases per profile.}
\label{tab:performance}
\scriptsize
\begin{tabular}{lrrr}
\toprule
\textbf{Profile} & \textbf{Min} & \textbf{Median} & \textbf{Max} \\
\midrule
\multicolumn{4}{l}{\textit{Compiler prototype}} \\
Accountability & 12.6 & 14.6 & 33.0 \\
Fairness & 55.3 & 58.7 & 270.4 \\
Combined & 59.5 & 61.1 & 100.3 \\
\midrule
\multicolumn{4}{l}{\textit{OKB prototype}} \\
EU (3 blocks) & 15.1 & 15.4 & 31.7 \\
US (2 blocks) & 13.7 & 14.9 & 19.0 \\
China (2 blocks) & 13.5 & 14.9 & 19.2 \\
EU+Fairness (4 blocks) & 61.6 & 61.6 & 248.1 \\
\bottomrule
\end{tabular}
\end{table}

Table~\ref{tab:performance}: All median latencies are within decision-time feasibility for second-scale allocation workloads. The Fairness maximum (270.4\,ms) exceeding Combined (100.3\,ms) reflects single-run variance in a small sample: medians (58.7 vs 61.1\,ms) are nearly identical, confirming B5Shape dominates both profiles with no systematic architectural difference.

\section{Discussion}

The central result is that governance reconfiguration is achieved entirely through profile selection, with no change to service code, evidence schema, or validation engine. Profile equivalence testing confirms Combined is strictly the most comprehensive profile: it subsumes both Accountability and Fairness, detecting all failure cases across both constituent profiles, with no equivalence pairs between any two profiles.

\emph{Compiler specification:} Each IR record specifies: \texttt{obligation\_id}, \texttt{target\_class}, \texttt{constraint\_type} (structural\,$|$\,sparql), \texttt{relation}, optional \texttt{threshold\_ref}, \texttt{severity}, and \texttt{message}. The compiler maps each record to a SHACL \texttt{NodeShape} deterministically: structural constraints become \texttt{sh:property} paths; SPARQL constraints become \texttt{sh:SPARQLConstraint}. Duplicate obligation IDs are rejected; severity merging follows Violation\,$>$\,Warning\,$>$\,Info. Full schema and worked examples are in the repository.

\emph{Scope:} The B5 fairness constraint is \emph{illustrative}: a two-group disparity proxy demonstrating the SHACL-SPARQL mechanism. Richer multi-group or temporal metrics are expressible within the same framework; selecting which to encode is a governance decision. Governance coverage claims should be understood accordingly. Companion works address collective governance evaluation~\cite{Sharma2025CollectiveFramework} and comparative regulatory analysis across the EU, US, and China~\cite{Sharma2026GlobalRegulation}.

\emph{Threats to validity:} \emph{Internal:} Evidence cases are hand-crafted; they may not cover all failure modes. An automated schema validation step would prevent silent target-class mismatches before deployment. \emph{External:} Evaluation uses a single domain (HPC scheduling); generalization to domains such as credit scoring or clinical AI requires further validation. \emph{Construct:} The fairness metric is a simplified proxy; the mechanism generalises, but metric selection is domain-specific.

\section{Conclusion}

Ontological Knowledge Blocks provide a four-layer pipeline: human-reviewed Regulatory IR, deterministic compilation into RDF/OWL schema and SHACL shapes, evidence graph emission by AI services, and profile-based validation. Three findings are central: (1)~profile sensitivity confirms identical evidence yields verdicts aligned with active obligations; (2)~profile equivalence testing confirms Combined is strictly the most comprehensive profile, subsuming both Accountability and Fairness with no equivalence pairs; (3)~violation accumulation is strictly additive across all cases. The illustrative B5 fairness constraint demonstrates the SHACL-SPARQL mechanism; richer multi-group metrics are expressible within the same framework. Validation latency of 12.6--100.3\,ms (median) demonstrates decision-time feasibility for second-scale allocation workloads. Future work targets automated schema validation to prevent silent constraint failures, high-throughput evaluation at $10^3$--$10^4$ decision scale, cross-domain validation (e.g., credit scoring, clinical AI), and richer multi-group fairness metrics.

\section*{Acknowledgments}
{\footnotesize\textit{\copyright~2026~IEEE. Personal use of this material is permitted. Permission from IEEE must be obtained for all other uses. Accepted at the Security, Trust and Privacy for Software and Applications (STPSA) Workshop, IEEE COMPSAC 2026, Madrid, Spain, July 7--10, 2026.}}

This work was supported by NHR at GWDG and conducted within the KISSKI project (BMBF grant no.\ 01\,IS\,22\,093\,A-E) at the University of Göttingen. All artefacts (SHACL shapes, IR records, evidence graphs, evaluation scripts) are released as open source at \url{https://github.com/AasishKumarSharma/open-knowledge-blocks}.

\appendices

\section{Evidence Graph Cases for Profile-Based Validation}
\label{app:cases}

This appendix provides the concrete RDF evidence graphs used in the evaluation. Each case is an \emph{evidence graph} describing one scheduling decision and its associated artefacts. The profiles in Section~\ref{sec:prototype} validate these graphs using SHACL: the Accountability profile targets provenance completeness (usage log and model artifact linkage); the Fairness/Transparency profile targets explanation presence and quantitative disparity; the Combined profile enforces both.

\subsection{Case A: Conformant Evidence}
Case~A (\texttt{case\_conform.ttl}) includes a usage log with timestamp and CPU time, a resolvable explanation URI, and a provenance chain in which the generating activity \texttt{prov:used} both a model artifact and a log artifact. Allocations of 120.0 and 110.0 GPU-hours yield a disparity of 0.083, within the 0.20 threshold. This case should conform under all profiles.

\begin{lstlisting}[caption={Case A (\texttt{case\_conform.ttl}): conformant evidence graph satisfying all obligation types.},label={lst:case_conform}]
@prefix ex:   <http://example.org/okb#> .
@prefix prov: <http://www.w3.org/ns/prov#> .
@prefix xsd:  <http://www.w3.org/2001/XMLSchema#> .

ex:svc1 a ex:AIService ;
  ex:hasDecision ex:dec1 .

ex:dec1 a ex:Decision ;
  ex:hasUsageLog ex:log1 ;
  ex:hasExplanation ex:exp1 ;
  ex:allocatedGPUHoursGroupA "120.0"^^xsd:decimal ;
  ex:allocatedGPUHoursGroupB "110.0"^^xsd:decimal ;
  ex:fairnessThreshold "0.20"^^xsd:decimal ;
  prov:wasGeneratedBy ex:act1 .

ex:log1 a ex:UsageLog ;
  ex:cpuTimeSeconds "3600.0"^^xsd:decimal ;
  ex:energyKWh "2.3"^^xsd:decimal ;
  ex:timestamp "2026-02-01T10:00:00Z"^^xsd:dateTime .

ex:exp1 a ex:Explanation ;
  ex:explanationURI "http://example.org/evidence/explanations/dec1"^^xsd:anyURI .

ex:act1 a prov:Activity, ex:Activity ;
  prov:used ex:model_v1, ex:log_art1 .

ex:model_v1 a ex:ModelArtifact .
ex:log_art1 a ex:LogArtifact .
\end{lstlisting}

\subsection{Case B: Missing Explanation}
Case~B (\texttt{case\_missing\_explanation.ttl}) contains full logging and provenance but omits the explanation artefact. It should fail profiles that enforce transparency (Fairness, Combined) and pass Accountability.

\begin{lstlisting}[caption={Case B (\texttt{case\_missing\_explanation.ttl}): missing explanation evidence.},label={lst:case_missing}]
@prefix ex:   <http://example.org/okb#> .
@prefix prov: <http://www.w3.org/ns/prov#> .
@prefix xsd:  <http://www.w3.org/2001/XMLSchema#> .

ex:svc1 a ex:AIService ;
  ex:hasDecision ex:dec2 .

ex:dec2 a ex:Decision ;
  ex:hasUsageLog ex:log2 ;
  ex:allocatedGPUHoursGroupA "120.0"^^xsd:decimal ;
  ex:allocatedGPUHoursGroupB "110.0"^^xsd:decimal ;
  ex:fairnessThreshold "0.20"^^xsd:decimal ;
  prov:wasGeneratedBy ex:act2 .

ex:log2 a ex:UsageLog ;
  ex:cpuTimeSeconds "3600.0"^^xsd:decimal ;
  ex:timestamp "2026-02-01T10:05:00Z"^^xsd:dateTime .

ex:act2 a prov:Activity, ex:Activity ;
  prov:used ex:model_v1 .

ex:model_v1 a ex:ModelArtifact .
\end{lstlisting}

\subsection{Case C: Missing Model Artifact}
Case~C (\texttt{case\_missing\_model\_artifact.ttl}) provides a provenance activity but does not link it to a model artefact via \texttt{prov:used}. It correctly fails under Accountability and Combined profiles: A5Shape targets \texttt{ex:Activity}, which is co-typed on all activity nodes alongside \texttt{prov:Activity}.

\begin{lstlisting}[caption={Case C (\texttt{case\_missing\_model\_artifact.ttl}): provenance incomplete (model artefact not referenced).},label={lst:case_model}]
@prefix ex:   <http://example.org/okb#> .
@prefix prov: <http://www.w3.org/ns/prov#> .
@prefix xsd:  <http://www.w3.org/2001/XMLSchema#> .

ex:svc1 a ex:AIService ;
  ex:hasDecision ex:dec3 .

ex:dec3 a ex:Decision ;
  ex:hasUsageLog ex:log3 ;
  ex:hasExplanation ex:exp3 ;
  ex:allocatedGPUHoursGroupA "120.0"^^xsd:decimal ;
  ex:allocatedGPUHoursGroupB "110.0"^^xsd:decimal ;
  ex:fairnessThreshold "0.20"^^xsd:decimal ;
  prov:wasGeneratedBy ex:act3 .

ex:log3 a ex:UsageLog ;
  ex:cpuTimeSeconds "3600.0"^^xsd:decimal ;
  ex:timestamp "2026-02-01T10:10:00Z"^^xsd:dateTime .

ex:exp3 a ex:Explanation ;
  ex:explanationURI "http://example.org/evidence/explanations/dec3"^^xsd:anyURI .

ex:act3 a prov:Activity, ex:Activity ;
  prov:used ex:log_art3 .

ex:log_art3 a ex:LogArtifact .
\end{lstlisting}

\subsection{Case D: Disparity Exceeds Threshold}
Case~D (\texttt{case\_disparity\_exceeds.ttl}) includes full logging, an explanation, and a correct provenance chain, but reports group allocations whose normalised disparity exceeds the declared threshold. It should fail Fairness and Combined profiles.

\begin{lstlisting}[caption={Case D (\texttt{case\_disparity\_exceeds.ttl}): quantitative fairness violation.},label={lst:case_disparity}]
@prefix ex:   <http://example.org/okb#> .
@prefix prov: <http://www.w3.org/ns/prov#> .
@prefix xsd:  <http://www.w3.org/2001/XMLSchema#> .

ex:svc1 a ex:AIService ;
  ex:hasDecision ex:dec4 .

ex:dec4 a ex:Decision ;
  ex:hasUsageLog ex:log4 ;
  ex:hasExplanation ex:exp4 ;
  ex:allocatedGPUHoursGroupA "200.0"^^xsd:decimal ;
  ex:allocatedGPUHoursGroupB "50.0"^^xsd:decimal ;
  ex:fairnessThreshold "0.20"^^xsd:decimal ;
  prov:wasGeneratedBy ex:act4 .

ex:log4 a ex:UsageLog ;
  ex:cpuTimeSeconds "3600.0"^^xsd:decimal ;
  ex:timestamp "2026-02-01T10:15:00Z"^^xsd:dateTime .

ex:exp4 a ex:Explanation ;
  ex:explanationURI "http://example.org/evidence/explanations/dec4"^^xsd:anyURI .

ex:act4 a prov:Activity, ex:Activity ;
  prov:used ex:model_v1 .

ex:model_v1 a ex:ModelArtifact .
\end{lstlisting}

\subsection{Expected Outcomes by Profile}
Table~\ref{tab:expected} summarises intended and observed conformance behaviour. All outcomes match expected behaviour.

\begin{table}[h]
\centering
\caption{Expected conformance outcome by profile and case. $\checkmark$\,=\,Pass; $\times$\,=\,Fail.}
\label{tab:expected}
\scriptsize
\begin{tabular}{p{11em}ccc}
\toprule
\textbf{Case} & \textbf{Acct.} & \textbf{Fair.} & \textbf{Combined} \\
\midrule
A: Conformant & $\checkmark$ & $\checkmark$ & $\checkmark$ \\
B: Missing explanation & $\checkmark$ & $\times$ & $\times$ \\
C: Missing model artefact & $\times$ & $\checkmark$ & $\times$ \\
D: Disparity exceeds & $\checkmark$ & $\times$ & $\times$ \\
\bottomrule
\end{tabular}
\end{table}

\end{document}